\documentclass[letterpaper, 10 pt, conference]{ieeeconf}  % Comment this line out if you need a4paper

\IEEEoverridecommandlockouts                              % This command is only needed if 
\usepackage{hyperref}
\usepackage{cite}
\usepackage{amsmath,amssymb,amsfonts}
\usepackage{graphicx}
\usepackage{caption}
\usepackage{textcomp}
\usepackage{xcolor}
\usepackage{booktabs}
\usepackage{graphicx}
\usepackage{url}
\usepackage{subfigure}
\usepackage{algpseudocode}
\usepackage{amsmath}
\usepackage{algorithm2e}
\usepackage{mwe}
\usepackage{balance}
\def\BibTeX{{\rm B\kern-.05em{\sc i\kern-.025em b}\kern-.08em
    T\kern-.1667em\lower.7ex\hbox{E}\kern-.125emX}}

\newcommand{\eg}{{\em e.g.,~}}
\newcommand{\ie}{{\em i.e.,~}}

\title{\LARGE \bf
Fast and Accurate Task Planning using Neuro-Symbolic Language Models and Multi-level Goal Decomposition
}

\author{Minseo Kwon, Yaesol Kim, and Young J. Kim% <-this % stops a space
\thanks{The authors are with the Department of Computer Science and Engineering at Ewha Womans University in Korea  ${\it \{minseo.kwon|kimy\}@ewha.ac.kr, kimyaesol@gmail.com}$.}
}

\begin{document}

\maketitle
\thispagestyle{empty}
\pagestyle{empty}

%%%%%%%%%%%%%%%%%%%%%%%%%%%%%%%%%%%%%%%%%%%%%%%%%%%%%%%%%%%%%%%%%%%%%%%%%%%%%%%%

\begin{abstract}
In robotic task planning, symbolic planners using rule-based representations like PDDL are effective but struggle with long-sequential tasks in complicated environments due to exponentially increasing search space. Meanwhile, LLM-based approaches, which are grounded in artificial neural networks, offer faster inference and commonsense reasoning but suffer from lower success rates.
To address the limitations of the current symbolic (slow speed) or LLM-based approaches (low accuracy), we propose a novel neuro-symbolic task planner that decomposes complex tasks into subgoals using LLM and carries out task planning for each subgoal using either symbolic or MCTS-based LLM planners, depending on the subgoal complexity. 
This decomposition reduces planning time and improves success rates by narrowing the search space and enabling LLMs to focus on more manageable tasks. Our method significantly reduces planning time while maintaining high success rates across task planning domains, as well as real-world and simulated robotics environments. 
More details are available at \href{http://graphics.ewha.ac.kr/LLMTAMP/}{http://graphics.ewha.ac.kr/LLMTAMP/}.
\end{abstract}

%%%%%%%%%%%%%%%%%%%%%%%%%%%%%%%%%%%%%%%%%%%%%%%%%%%%%%%%%%%%%%%%%%%%%%%%%%%%%%%%

%%%%%%%%%%%%%%%%%%%%%%%%%%%%%%%%%%%%%%%%%%%%%%%%%%%%%%%%%%%%%%%%%%%%%%%%%%%%%%%%

\section{INTRODUCTION}

In the field of AI planning, symbolic language-based planning using logic formulations 
such as Planning Domain Definition Language (PDDL) \cite{fox2003pddl2} has been effective in generating valid plans across various domains. Such use of symbolic language in robotic task planning is traced back to the Shakey robot project in the early 1970s using STRIPS\cite{lavalle:2006}. 
However, since the time complexity of these symbolic planners is known to be PSPACE-hard\cite{helmert2006fast}, solving long-sequential tasks in domains with extensive search spaces using these symbolic planners is intractable, making their practical application to robot task planning limited. Recently, Large Language Models (LLMs) have shown advantages as autonomous robot task planners due to the short inference time compared to symbolic planners and their ability to leverage commonsense knowledge and generalization capabilities\cite{zhao2024large}.  

At a high level, the use of LLMs for task planning is divided into treating LLMs as a policy model (known as {\em L-Policy}) or as a world model (known as {\em L-Model})\cite{zhao2024large}. L-Policy exploits the commonsense knowledge of LLMs to directly query proper policy for a given state, while L-Model utilizes LLMs as a simulation model of the world to query the state of the world as a result of executing an action or a policy.
However, despite their strengths, LLMs suffer from token inefficiency and correction inefficiency \cite{hu2023tree}, often generating hallucinated action sequences and failing on complex tasks \cite{valmeekam2023planning}. 
To address the limitations of current LLM-based task planners, we propose a novel neuro-symbolic task planner that leverages LLMs as both L-Policy and L-Model to solve a long-sequential robotic task. Our planner is significantly faster than symbolic planners and more accurate than LLM-based planners.

An immediate issue in handling long-sequential tasks using LLMs is LLM's token inefficiency since the planning descriptions involve a long and repetitive sequence of world and robotic states and a history of policies and their results. To circumvent this issue, we utilize LLMs as L-Model to generate a sequence of subgoals for a long-horizon task, effectively decomposing it into smaller and manageable sub-tasks. This goal decomposition also provides a useful side-effect to reduce the overall search space, yielding an accurate subgoal planner based on LLMs. Indeed, we use the Monte Carlo Tree Search (MCTS) algorithm while using LLMs as L-Policy to accurately solve each subgoal, reducing the correction inefficiency common in LLM-based planners. 

Furthermore, if the original task is moderately complex, requiring a smaller minimum description length (MDL)\cite{zhao2024large} to solve the given problem, one can rely on a symbolic planner to solve the subgoals precisely while effectively avoiding the exponential growth of planning time.

Overall, our planning pipeline consists of three major steps:
\begin{enumerate}
\item {\bf Planning formulation:} Given a planning goal in natural language description and domain knowledge, our task planner relies on PDDL to encode the problem descriptions. We also obtain the semantic and spatial relationships of target objects in the environment using a multimodal LLM, translated and encoded in problem PDDL.

\item {\bf Subgoal generation:} We utilize the L-Model to generate a sequence of subgoals by decomposing the given goal.

\item {\bf Task planning:} If the MDL is moderate, we rely on a symbolic planner to solve each subgoal; otherwise, we generate and expand a search tree and use the MCTS algorithm with L-Policy as a rollout policy to solve the subgoal. This subgoal planning is repeated for each sub-task, and the plans are combined to form the overall plan.
\end{enumerate}

We conducted experiments across three task planning domains while varying the problem complexity. Compared to the state-of-the-art symbolic task planner like the Fast Downward planner \cite{helmert2006fast}, our approach significantly reduced planning time while maintaining an acceptable success rate. Additionally, we conducted experiments using dual robot manipulators and a robotic simulator to demonstrate the practical utility of our planner.

In summary, the main contributions of our work are:
\begin{itemize}
    \item We propose a novel neuro-symbolic task planning pipeline for executing complex robotic tasks on physical robots utilizing LLMs as both L-Model and L-Policy. 
    
    \item L-Model is used to decompose the given goal into multi-level subgoals to reduce the planning time while increasing the planning success rates. L-Policy is exploited to plan subgoals combined with MCTS. For a moderately complex planning task, a symbolic planner is alternatively used to guarantee more accurate planning results.

    \item Experimentally, we have shown that our new planner achieves an average success rate of $88.2\%\sim 100\%$  while the planning time is only $3.3\times \sim 10.2\times$ slower than the baseline LLM planner, which approaches zero success rate, depending on the problem complexity. 

    \item We demonstrate the applicability of our new planner on both real and simulated robot task planning scenarios. We also perform an ablation study to demonstrate the effectiveness of our goal decomposition strategy.
\end{itemize}

The rest of this paper is organized as follows. In Sec.~\ref{sec:prev}, we review relevant work to task planning. In Sec.~\ref{sec:pipeline}, we outline the overall pipeline, and in Sec.~\ref{sec:subgoal planner}, we explain the algorithms of both the symbolic subgoal planner and the LLM-based subgoal planner. In Sec.~\ref{sec:experiments}, we present the task planning results and experiments in real and simulation robotics environments and conclude the paper and discuss future work in Sec.~\ref{sec:conclusion}.

%%%%%%%%%%%%%%%%%%%%%%%%%%%%%%%%%%%%%%%%%%%%%%%%%%%%%%%%%%%%%%%%%%%%%%%%%%%%%%%%

%%%%%%%%%%%%%%%%%%%%%%%%%%%%%%%%%%%%%%%%%%%%%%%%%%%%%%%%%%%%%%%%%%%%%%%%%%%%%%%%
\section{RELATED WORK}\label{sec:prev}

\subsection{Symbolic Robot Task Planning}
Symbolic or rule-based robot task planning is rooted in classical AI planning using symbolic languages and has been extensively studied for over four decades\cite{lavalle:2006}. We refer the readers to recent surveys on this topic, such as \cite{Guo2023RecentTI}. The current trend in symbolic task planning is to use hierarchical planning to solve a complex problem or to integrate it with geometric motion planning, known as Task and Motion Planning (TAMP) \cite{garrett2021integrated}. However, the intrinsically high time complexity of symbolic planning hinders its scalability to adapt to the physical world \cite{lavalle:2006}.

\subsection{LLM-based Robot Task Planning}
Recent studies have explored using LLMs for robot task planning by leveraging their real-world understanding. \cite{ahn2022can} combines language understanding with action grounding in real-world affordances, enabling robots to execute tasks based on their capabilities. Similarly, \cite{singh2023progprompt} introduced a prompting scheme that enables LLM to generate Python codes composed of robot action primitives, incorporating environmental state feedback. 
% 수정
\cite{gao2024physically} fine-tuned multimodal LLMs to integrate physical grounding with visual inputs for task planning.
TAMP has also been addressed using LLM by \cite{ding2023task}, enabling LLM as spatial relationship generators between environment objects. 
Additionally, some studies combined LLM-based high-level planning with reinforcement learning for low-level control \cite{dalal2024plan}.
However, these approaches' common limitations are low success rates in solving long sequential tasks, limited multi-step reasoning, and weak failure recovery.

% 수정
Recent Large Reasoning Models have shown high success rates across various PDDL domains without additional frameworks, their performance still declines as task complexity increases, with success rates approaching zero in highly constrained domains \cite{wang2024planning}.

\subsection{Hybird Task Planning}
Recently, studies have been conducted on integrating LLMs with symbolic planning methods. \cite{liu2023llm+} and \cite{xie2023translating} used LLMs to translate natural language problem descriptions into PDDL initial states and goals through few-shot prompting. However, these approaches struggle in real-world applications where problems are not presented in natural language.
\cite{shirai2023vision} combined LLMs with vision models to generate planning problem specifications based on real-world scenes, using re-prompting to correct specification errors. 
LLMs have also been used to solve PDDL problems. \cite{silver2022pddl} showed that while LLMs can solve some non-trivial PDDL problems, they often fail on more complex tasks, though their outputs can guide heuristic planners. 
% Building on this, \cite{silver2024generalized} proposed a method where LLMs generate Python functions to create PDDL plans with automated debugging. \cite{zhou2024isr} introduced a framework that iteratively refines PDDL plans using validator feedback. While these methods have improved success rates compared to LLM-only methods, they have been tested mostly on small-size problems. 
\cite{silver2024generalized} improved this by generating Python functions for PDDL planning with automated debugging, while \cite{zhou2024isr} introduced an iterative refinement framework using validator feedback. While these methods have improved success rates compared to LLM-only methods, they have been tested mostly on small-scale problems. 

\subsection{Integrating LLMs with Tree Search}
% Combining tree structures with LLM-generated actions has been explored. \cite{yao2024tree} samples possible next actions from the current state using an LLM and selects the best action via an LLM-evaluator, repeating this process with DFS or BFS. 
% To address this method's token and runtime inefficiencies, \cite{hu2023tree} proposes sampling multiple plans at once rather than repeatedly calling the LLM to generate an action tree and proposes selecting actions from the tree based on observations and histories.
% \cite{hao2023reasoning} integrates MCTS with LLMs by iteratively calculating state transitions in MDPs, and \cite{zhao2024large} also uses LLMs as both \textit{L-Model} and \textit{L-Policy} combined with MCTS to solve large-scale POMDPs. 

% Our method also uses an LLM to sample multiple plans, and exploit MCTS to find a plan. Still, it differs from \cite{hu2023tree} in that ours relies on LLM-induced goal decomposition to generate multiple subplans and solve a deterministic problem, unlike \cite{zhao2024large}. Our MCTS is performed on a fixed tree for a sub-problem, not the entire planning problem.

Combining tree structures with LLM-generated actions has been explored in various studies. \cite{yao2024tree} samples possible next actions from the current state using an LLM and selects the best one via an LLM-evaluator, iterating with DFS or BFS. To improve token and runtime efficiency, \cite{hu2023tree} proposes sampling multiple plans at once to generate an action tree and selecting actions from the tree based on observations and histories. 
\cite{hao2023reasoning} integrates MCTS with LLMs for iterative state transitions in MDPs, and \cite{zhao2024large} employs LLMs as both L-Model and L-Policy within MCTS to solve large-scale POMDPs.

Our method also samples multiple plans at once using an LLM and applies MCTS, but unlike \cite{hu2023tree}, ours relies on LLM-induced goal decomposition to generate multiple deterministic sub-problems. Unlike \cite{zhao2024large}, our MCTS operates on a fixed tree for a sub-problem rather than the entire planning problem.

%%%%%%%%%%%%%%%%%%%%%%%%%%%%%%%%%%%%%%%%%%%%%%%%%%%%%%%%%%%%%%%%%%%%%%%%%%%%%%%%

%%%%%%%%%%%%%%%%%%%%%%%%%%%%%%%%%%%%%%%%%%%%%%%%%%%%%%%%%%%%%%%%%%%%%%%%%%%%%%%%

\section{TASK PLANNING PIPELINE} \label{sec:pipeline}

\begin{figure*}
    \centering
    \includegraphics[width=\textwidth]{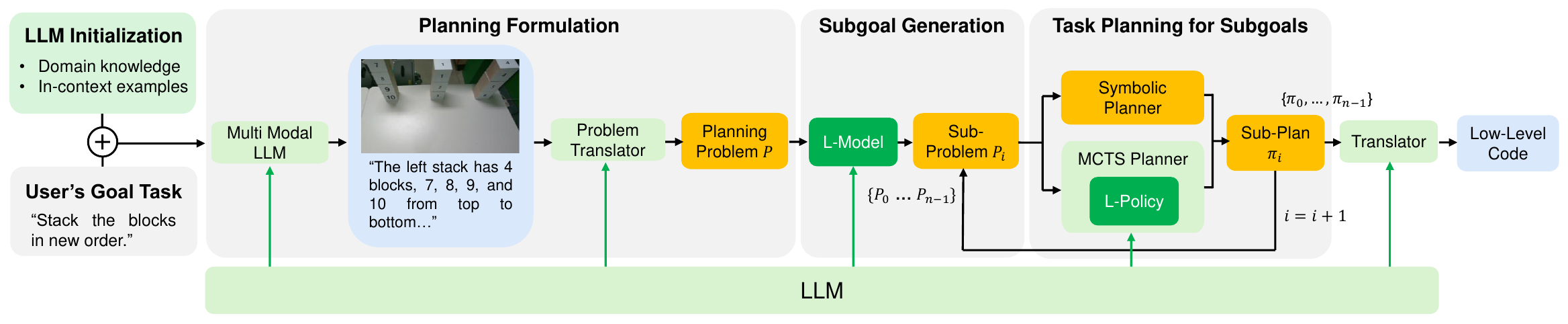}
    \caption{Neuro-symbolic task planning pipeline. LLM (the green blocks) and symbolic languages (the orange blocks) are used for various steps in the pipeline.
    }
    \label{fig1}
    \vspace{-1.0em}
\end{figure*}

We formulate our task planning problem as a multi-valued planning task (MPT)\cite{helmert2006fast} using a tuple: 
\begin{equation}\label{eq:tpform}
P \equiv \langle \mathcal{S}, \mathcal{O}, \mathcal{A}, \mathcal{T}, s_0, S^\star \rangle,
\end{equation}
where $\mathcal{S}$ is a finite set of fully observable states,$\mathcal{O}$ is environment objects, $\mathcal{A}$ is a finite set of possible actions, $\mathcal{T}: \mathcal{S}\times\mathcal{A} \rightarrow \mathcal{S}$ is a deterministic state transition function, $s_0 \in \mathcal{S}$ is an initial state, and $S^\star \subset \mathcal{S}$ is a set of goal states. Our planning objective is to find a policy $\pi = \{a_1, \cdots, a_n | \forall a_i\in \mathcal{A} \}$ for $P$ in Eq.~\ref{eq:tpform} to transit from $s_0$ to $\exists s_n \in S^\star$ in finite steps. Now, we explain each step in our planning pipeline to find a valid $\pi$ for $P$ and provide a more detailed explanation of the subgoal planner in the next section. An overview of our pipeline is also illustrated in Fig.~\ref{fig1}.

\subsection{Planning Formulation}\label{sec:form}
For the robot to fully understand and interact with its environment, both semantics and geometry about the objects in the environment are required.
We use a multimodal LLM such as GPT-4o\footnote{\url{https://openai.com}} to process image and text prompts simultaneously. By providing a color image captured by an RGBD camera along with the prompt, \eg \textit{"What objects are on the table? Tell me each of their appearance and spatial relationships."}, the LLM can describe the objects on the table, including their spatial relationships, positions, and appearance.
Given the scene description, user-provided goal task, the domain PDDL, and an in-context example, the LLM generates a problem PDDL consisting of environment objects $\mathcal{O}$, the initial state $s_0$, and the goal state $S^\star$ to specify the planning problem $P$. We utilize one-shot prompting \cite{brown2020language} by providing an example of problem PDDL generation to enhance the LLM's responses. 

We also employ a 2D open-vocabulary object detection model \cite{ren2024grounded} to estimate the geometric information, specifically the bounding box of the target objects identified by the multimodal LLM. These bounding boxes are essential for a robot manipulator to motion-plan their grasp poses.

\subsection{Subgoal Generation}\label{subgoal}
Solving a complex task by breaking it down into smaller, easier tasks is often effective\cite{ghallab2004automated}. 
In our case, while LLMs can directly generate relatively accurate plans for smaller tasks, their performance significantly decreases as the task complexity increases and the plan grows beyond a certain size \cite{valmeekam2023planning}.
To address this problem, we leverage the commonsense knowledge of LLMs, \ie the L-Model, to decompose a given goal into multiple subgoals, simplifying the planning process. 

Let us call an ordered set of $\mathcal{G} = \{S^\star_0, S^\star_1, \cdots, S^\star_{n}\}$ a {\em sequence of subgoals} or simply {\em subgoals} of $P$ in Eq.~\ref{eq:tpform} iff  $S^\star_i$ is {\em reachable} from $S^\star_{i-1}$ for $1\le \forall i \le n$ via a finite number of state transitions from $\exists s_{i-1} \in S^\star_{i-1}$ to $\exists s_{i} \in S^\star_{i}$   and $S^\star_0 = \{s_0\}, S^\star_n = S^\star$. Our objective is to decompose the original task problem $P$ into $n$ smaller sub-problems $P_i$'s,  $0\le \forall i \le n - 1$ as 
\begin{equation}\label{eq:subprob}
P_i \equiv \langle \mathcal{S}, \mathcal{O}, \mathcal{A}, \mathcal{T}, s_i, S^\star_{i+1} \rangle.
\end{equation}

We prompt the LLM with domain knowledge and a one-shot planning example along with the explanation of the steps for solving the problem and then ask the LLM to generate $\mathcal{G}$ by observing how the example problem is solved. 
For instance, in the Blocksworld-new domain, if the blocks are stacked in the order \texttt{(on b1 b2)(on b2 b3)(on-table b3 t1)}, the reverse order stacking requires each of the three blocks to be unstacked with no objects on each block—\texttt{(clear b1)(clear b2)(clear b3)(clear-table t1)}—to rearrange them appropriately.

\subsection{Task Planning}\label{task planning}

Once the subgoals $\mathcal{G}$ are generated, we attempt to find a policy $\pi_{i} \subset \pi$ for each sub-planning problem $P_i$. The role of the subgoal planner is explained in detail in Sec.~\ref{sec:subgoal planner}.
By sequentially applying actions from the policy $\pi_i$ to the initial state $s_i$, we determine the resulting state $s_{i+1}$. If $s_{i+1} \in S^\star_{i+1}$, $\pi_{i}$ is called a {\em valid} policy for $P_{i}$, and $s_{i+1}$ becomes the initial state for the next sub-problem $P_{i+1}$. 
Finally, by aggregating each valid policy $\pi_{0}, \pi_{1}, \dots, \pi_{n-1}$ for each sub-problem, we can obtain the final policy, $\pi=\bigcup_i \pi_i$, which is symbolically represented as a plan PDDL. LLM then translates the plan PDDL into robot-executable low-level code. The robot then automatically executes the corresponding actions by invoking predefined high-level robot action primitives, \eg such as \texttt{pick}, \texttt{place}\cite{vemprala2024chatgpt}.

%%%%%%%%%%%%%%%%%%%%%%%%%%%%%%%%%%%%%%%%%%%%%%%%%%%%%%%%%%%%%%%%%%%%%%%%%%%%%%%%

%%%%%%%%%%%%%%%%%%%%%%%%%%%%%%%%%%%%%%%%%%%%%%%%%%%%%%%%%%%%%%%%%%%%%%%%%%%%%%%%
\section{SUBGOAL PLANNER} \label{sec:subgoal planner}

\begin{figure*}
    \centering
    \includegraphics[width=\textwidth]{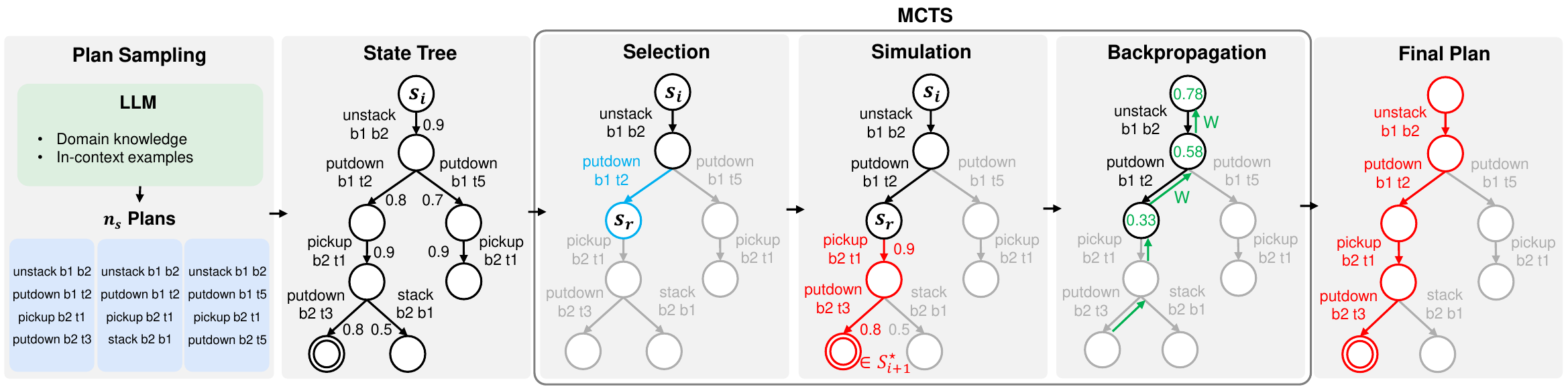}
    \caption{An overview of the MCTS LLM Planner. First, the L-Policy samples $n_s$ plans for a sub-problem $P_i$. For instance, the initial state $s_i$ of $P_i$ is \texttt{(on b1 b2)(on-table b2 t1)}, etc., and the goal state $S^\star_{i+1}$ should satisfy \texttt{(clear b1)(clear b2)(clear-table t1)}. 
    A state tree $T_i$ is then generated, and our MCTS algorithm uses $T_i$ to search for a plan that reaches $S^\star_{i+1}$.}
    \label{fig2}
    \vspace{-1.0em}
\end{figure*}

We employ two approaches to solve each sub-problem $P_i$ in Eq.~\ref{eq:subprob} using a symbolic planner or MCTS LLM planner depending on the size of $P_i$. We explain each of these planners.

\subsection{Symbolic LLM Planner}\label{sec.4.A}

When the size $|P_i|$ of $P_i$ is moderate, it is possible to use a symbolic planner rather than an LLM-based planner to solve $P_i$ precisely. However, estimating $|P_i|$ is not easy. In theory, one can use a problem measure like MDL \cite{zhao2024large} to estimate it, but in practice, deriving the MDL for a challenging task is quite hard. Instead, one may estimate an MDL-like metric for $P_i$ by empirically measuring the planning time spent by running the MCTS LLM planner or a symbolic planner for a sampled $P_i$. If such an estimate is sufficiently high, we assume that $P_i$ is complex and resort to the MCTS LLM planner in the next section; otherwise, we use a symbolic planner.

To solve $P_i$ symbolically, one can use any symbolic planner, but we opted for the Fast Downward planner\cite{helmert2006fast}, one of the fastest symbolic planners. This guarantees an exact solution to $P_i$ if one exists.

\subsection{MCTS LLM Planner}\label{sec.4.B}
When $|P_i|$ is high, using a symbolic planner to solve $P_i$ is impractical due to the high combinatorial search space. In this case, we use an MCTS planner combined with the LLM. 
As illustrated in Fig.~\ref{fig2}, our MCTS LLM planner first samples $n_s$ plans for a sub-problem $P_i$ using an LLM (\ie L-Policy), then building a state tree with the LLM-sampled plans, which serves as the reduced search space. The MCTS algorithm then searches this tree to identify an action sequence (\ie a policy) that leads to a state satisfying the subgoal $S^\star_{i+1}$.

\subsubsection{Plan Sampling}\label{plansampling}
Given the domain PDDL in Sec.~\ref{sec:form} and a few in-context planning examples, the LLM generates $n_s$ plans, $\{\pi^1_{i}, \pi^2_{i}, \cdots, \pi^{n_s}_i\}$ to achieve the subgoal in $P_i$. Unlike \cite{hu2023tree}, which samples the entire problem $P$, we sample only for a sub-problem $P_i$, leading to presumably higher accuracy.
Also, the {\em action weight} is computed as the sum of token log probabilities for each LLM-generated action, reflecting the LLM's confidence when generating the action \cite{xiong2023can}. Since a token's log probability represents its conditional probability given previous tokens, the action weight can be viewed as the conditional probability of the current action occurring, given the history of previous actions.
This action weight will guide the rollout process in the MCTS.

\subsubsection{State Tree Generation}\label{statetree}
We generate a state tree $T_i$ for $P_{i}$ by coalescing the sampled $n_s$ plans where each node in $T_i$ represents a state $s\in \mathcal{S}$ and each edge corresponds to an action $a\in \mathcal{A}$ connecting $s, s'$ when $s' = \mathcal{T}(s, a)$. $T_i$ bounds the MCTS search space, ensuring the search is restricted to valid LLM-generated actions.
Moreover, we verify whether the preconditions of $a$, as defined in the domain PDDL, hold for $s$. If valid, then $a$ is added to $T_i$; otherwise, subsequent actions are removed from $T_i$. 
This post-validity check is applied to every action in all sampled plans.

% experiment graphs

\begin{figure*}[htb]
    \centering
    \includegraphics[width=0.9\textwidth]{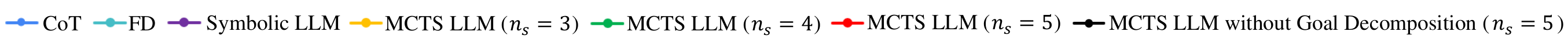}
    \label{legends_sr}
    \vspace{-1em}
\end{figure*}

\begin{figure*}[htb]
\centering
%\subfigure[Barman-new]{
    \includegraphics[width=0.31\textwidth]{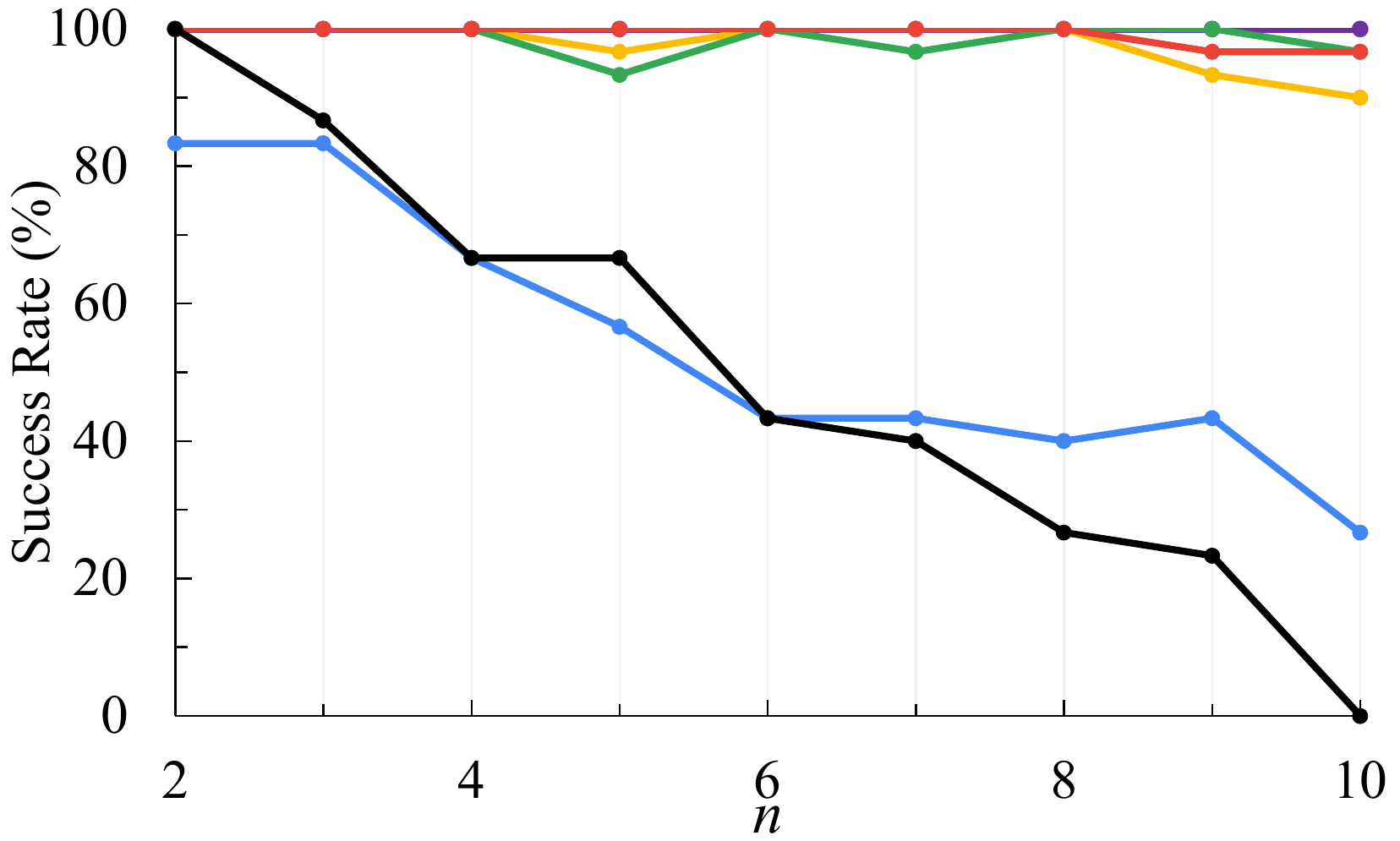}
%    \label{sr1}
%}
\hspace*{0.1em}
%\subfigure[Blocksworld-new]{
    \includegraphics[width=0.31\textwidth]{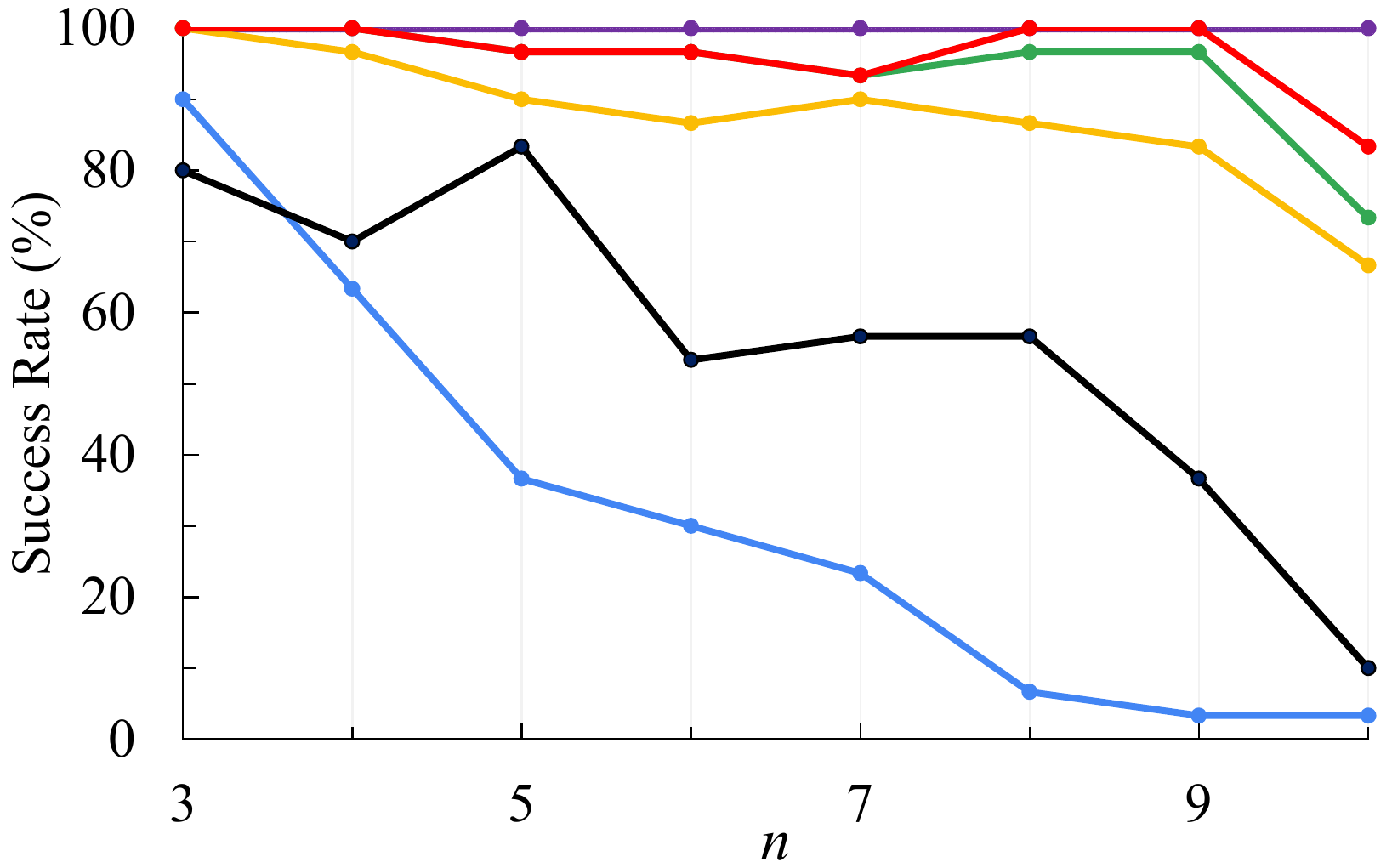}
%    \label{sr2}
%}
\hspace*{0.1em}
%\subfigure[Grippers-new]{
    \includegraphics[width=0.31\textwidth]{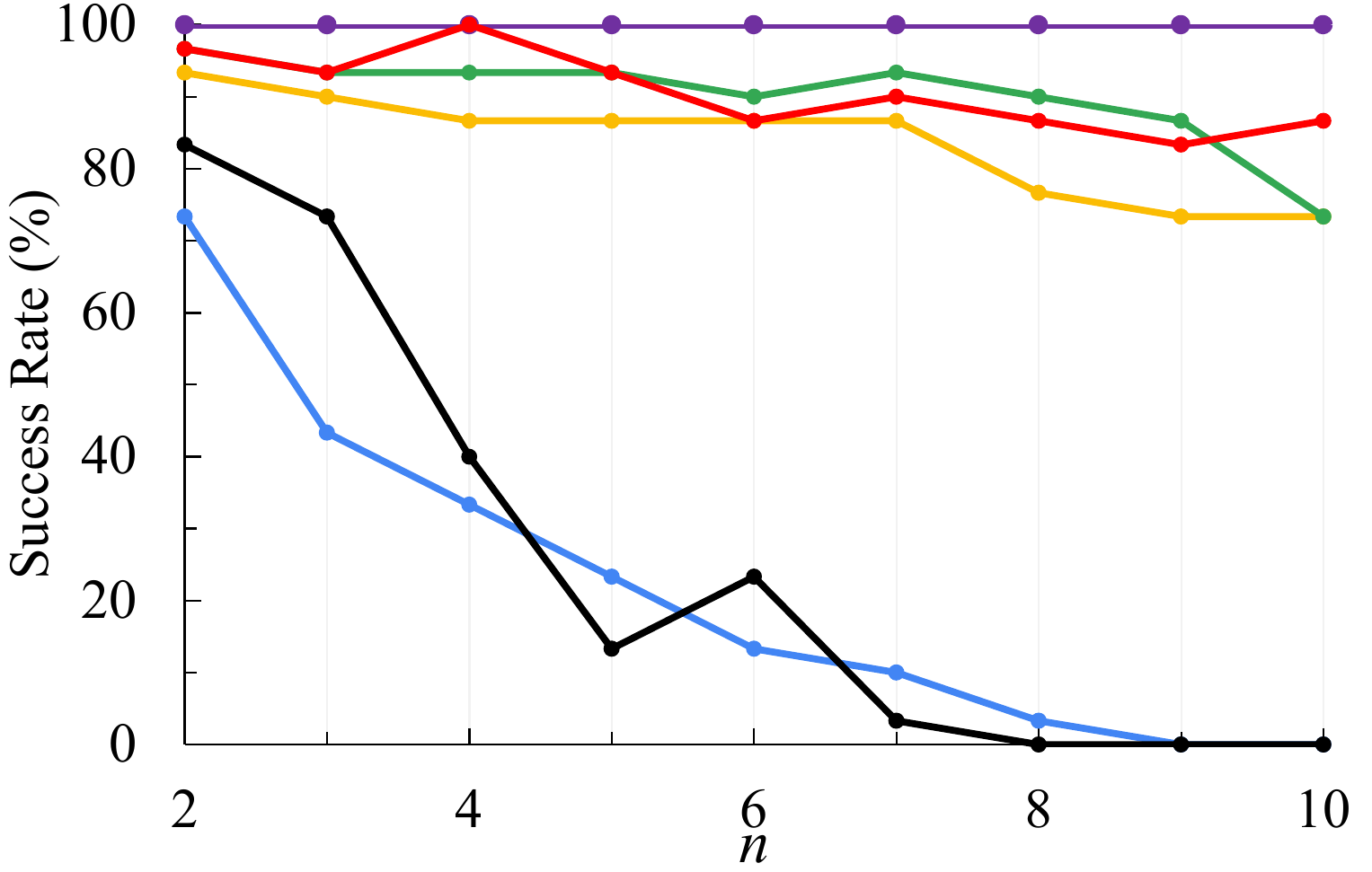}
%    \label{sr3}
%}
\subfigure[Barman-new]{
    \includegraphics[width=0.31\textwidth]{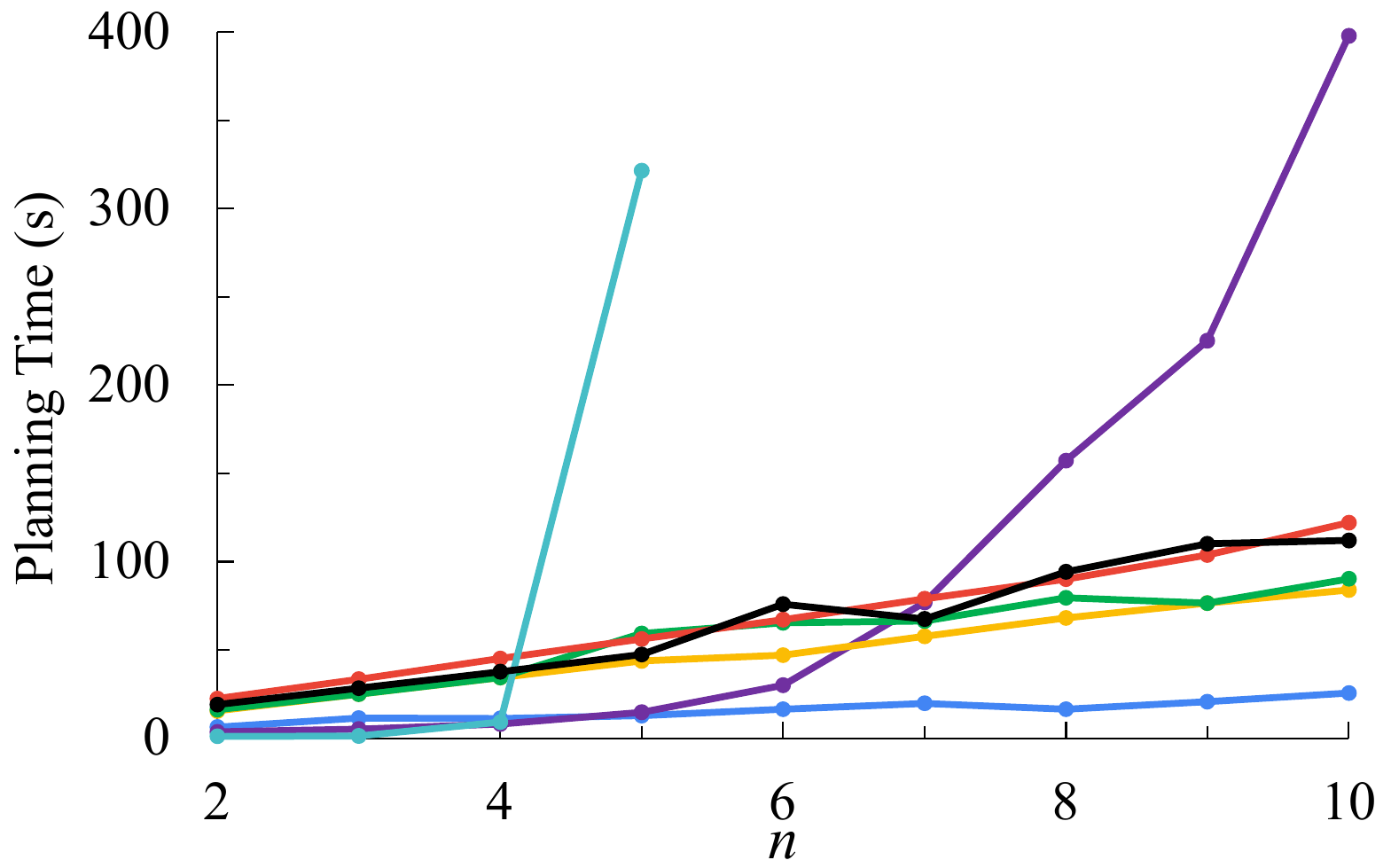}
%    \label{time1}
}
\subfigure[Blocksworld-new]{
    \includegraphics[width=0.31\textwidth]{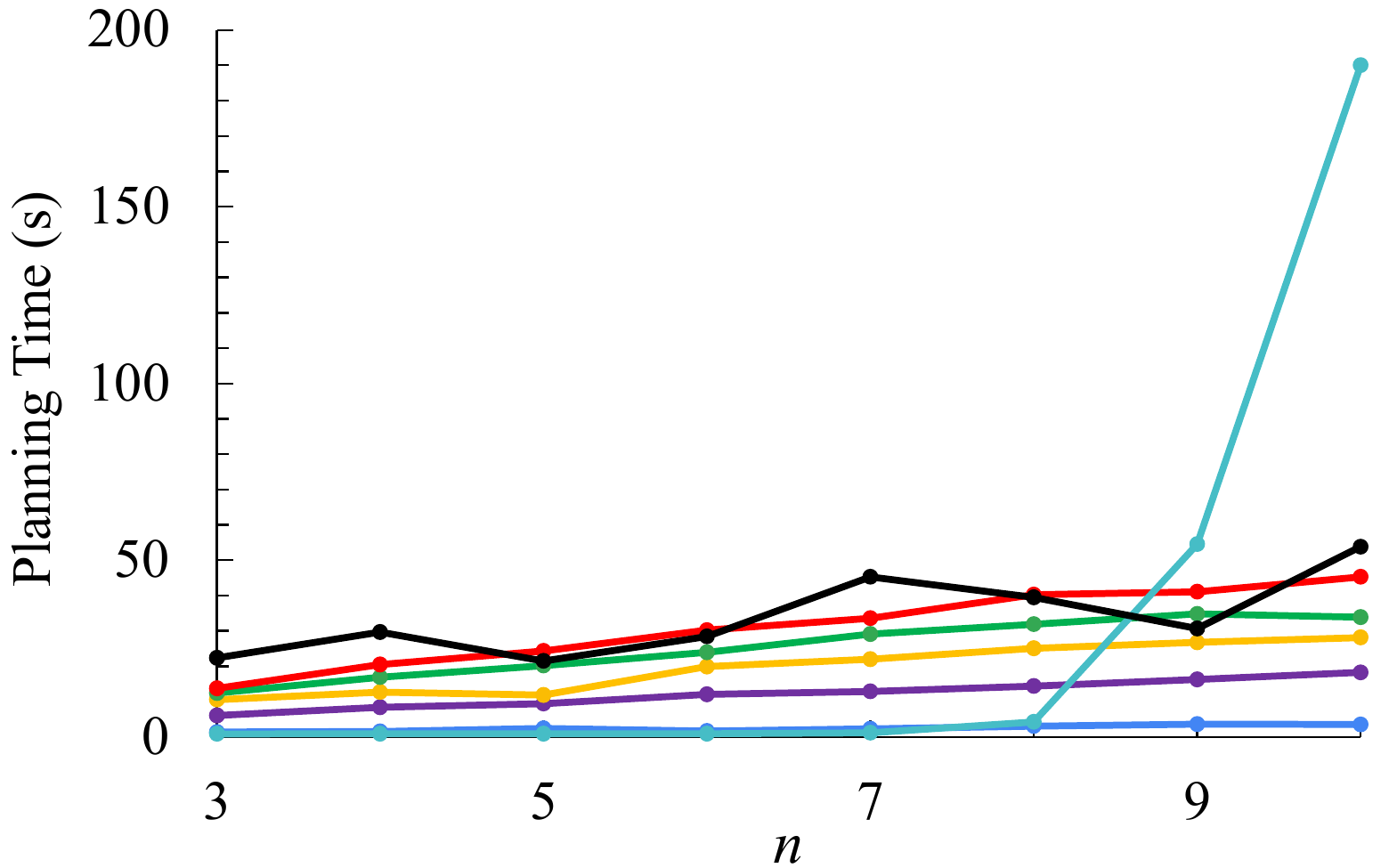}
%    \label{time2}
}
\subfigure[Gripper-new]{
    \includegraphics[width=0.31\textwidth]{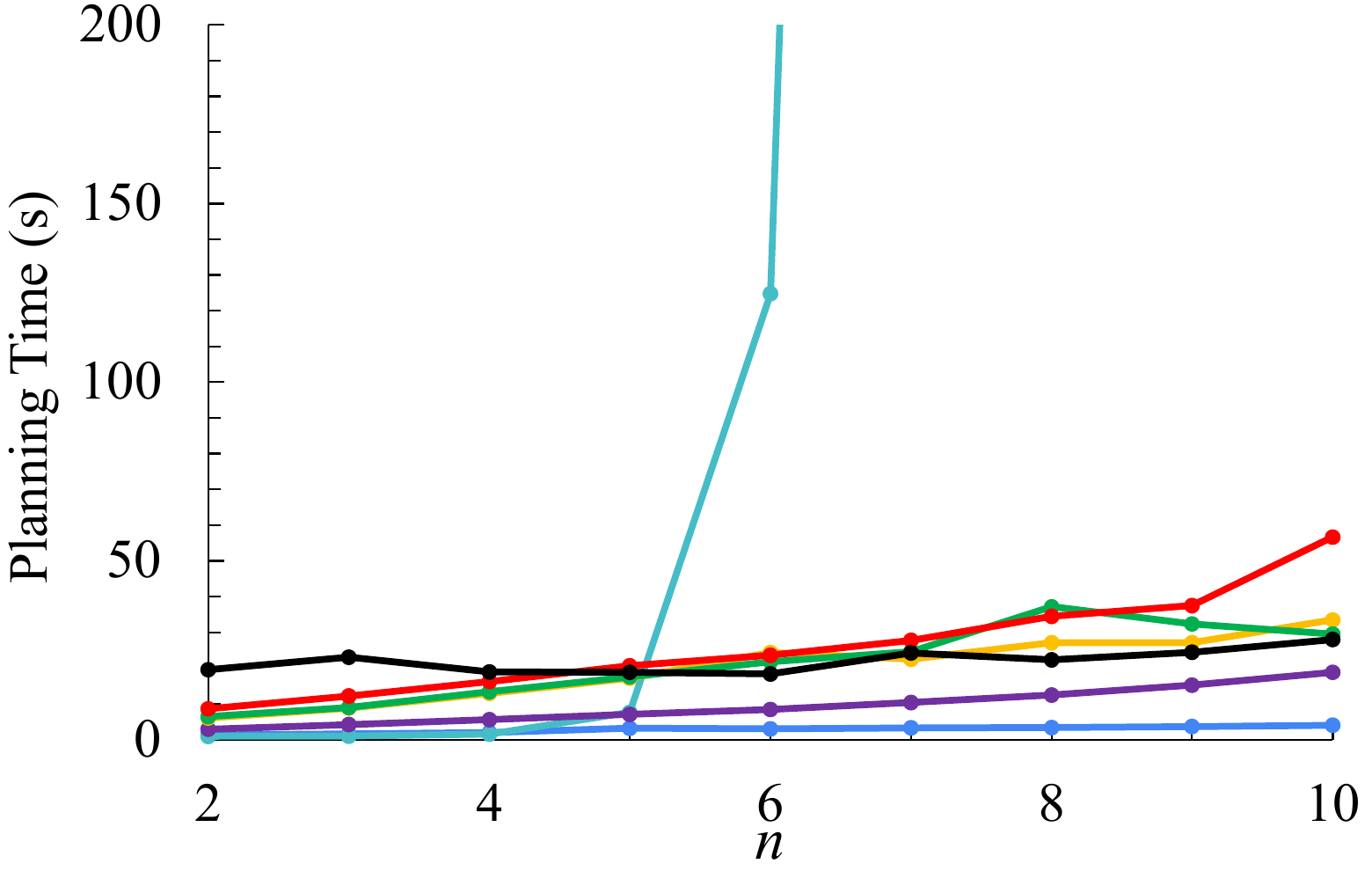}
%    \label{time3}
}
% \vspace{-0.7em}
\caption{Success rates (top row) and planning time (bottom row) of CoT, FD, Symbolic LLM, MCTS LLM planners with $3\le n_s \le 5$, and MCTS LLM planner without goal decomposition with $n_s=5$. The $x$ axis in all the graphs denotes the domain complexity $n$.
}
\label{fig.3}
\vspace{-1.5em}
\end{figure*}

\subsubsection{Monte Carlo Tree Search}
We search the state tree $T_i$ using the MCTS to find a policy $\pi_i$ for $P_i$. Our MCTS is quite different from conventional MCTS like \cite{hao2023reasoning} in that: 1)
we already expanded the tree $T_i$ that is fixed and constrains the overall search space, so the expansion step is not needed during the search; 
2) our rollout policy searches only within $T_i$. 
The goal of our MCTS is to estimate the reward for tree nodes and find a valid $\pi_i$ from the initial state $s_i$ to a goal state $s^\star \in S^\star_{i+1}$, guided by the rewards. The following selection, simulation, and backpropagation processes are repeated to find $\pi_i$.

{\bf 1. Selection:} % 수정: parameter 추가
Starting from the root node $s_i$, we recursively traverse $T_i$ by selecting the child node with the highest UCB1 score\cite{auer2002finite} (with an exploration parameter of 1) from the set of visited nodes. This continues until reaching a node whose all child nodes are visited for the first time. Then, one of the child nodes is randomly selected, say $s_r$.
If $s_r$ is included in the goal states  $s_r \in S^\star_{i+1}$, 
the MCTS stops immediately and $s_r$ is traced back to $s_i$, thereby constructing a plan $\pi_{i}$ for $P_{i}$.

{\bf 2. Simulation:}
The simulation step is rolled out and estimates the reward of $s_r$ passed from the selection process.
Our rollout policy works as follows:
among the possible next nodes (states) that can be transited from the current node (state), the node with the highest action weight (on the red edges in Fig.\ref{fig2}), already computed during the plan sampling step, is selected for the next node to visit.  
This process is repeated on the tree $T_i$ until a leaf node $s^\star$ is reached.
If $s^\star \in S^\star_{i+1}$, the returned reward is $\frac{1}{1 + d}$ where $d$ is the nodal distance from $s_r$ to $s^\star$; otherwise, zero reward is returned. If $s_r \in S^\star_{i+1}$, the reward is 1.

{\bf 3. Backpropagation:} 
The reward (the green nodal values in Fig.~\ref{fig2}) obtained from the simulation step is backpropagated to update the nodes traversed earlier, incrementing its visit count and adding the reward.

%%%%%%%%%%%%%%%%%%%%%%%%%%%%%%%%%%%%%%%%%%%%%%%%%%%%%%%%%%%%%%%%%%%%%%%%%%%%%%%%

%%%%%%%%%%%%%%%%%%%%%%%%%%%%%%%%%%%%%%%%%%%%%%%%%%%%%%%%%%%%%%%%%%%%%%%%%%%%%%%%

\section{EXPERIMENTS} \label{sec:experiments}

% 수정: parameter 추가
\subsection{Experimental Setup}
All task planning experiments were conducted on an Intel Core i9 CPU and NVIDIA RTX 6000 GPUs. We employed GPT-4o for the multimodal LLM with temperature 0.0 and used the Fast Downward planner as the symbolic PDDL planner. We conducted PDDL task planning experiments in three well-known IPC domains by modifying their problem complexities\cite{seipp-et-al-zenodo2022}: Barman-new, Blocksworld-new, and Gripper-new.

\paragraph{Barman-new}
This domain involves a dual-arm manipulator making cocktails. The goal is to prepare $2\le n \le 10$ cocktails, and each poured into a different shot glass, similar to examples in \cite{liu2023llm+}. The number of ingredients is three, and the number of shot glasses is $n+1$. 

\paragraph{Blocksworld-new}
In this domain, a robotic arm stacks $3\le n \le 10$ blocks, randomly divided into one to three stacks arranged on a table. The goal is to rearrange the blocks for each stack. 
Unlike the original Blocksworld domain, we increase the planning complexity by creating six block placement positions for the interim workspace. As a result, the planner must also specify positions for placing the blocks rather than using a single \texttt{on-table} predicate as in the original domain.

\paragraph{Gripper-new} 
In the Gripper-new domain, four robots move $2\le n \le 10$ balls to four different rooms from their initial location. We incorporate four multiple robots, making the planning process more complex in a multi-agent scenario, similar to \cite{liu2023llm+}. The positions of the balls and robots in both the initial and goal states are random.

For each $n$ in the above domains, we randomly generated 30 problem PDDL files for the experiments and measured the planning performances.

\begin{figure*}
    \centering
    \includegraphics[width=\textwidth]{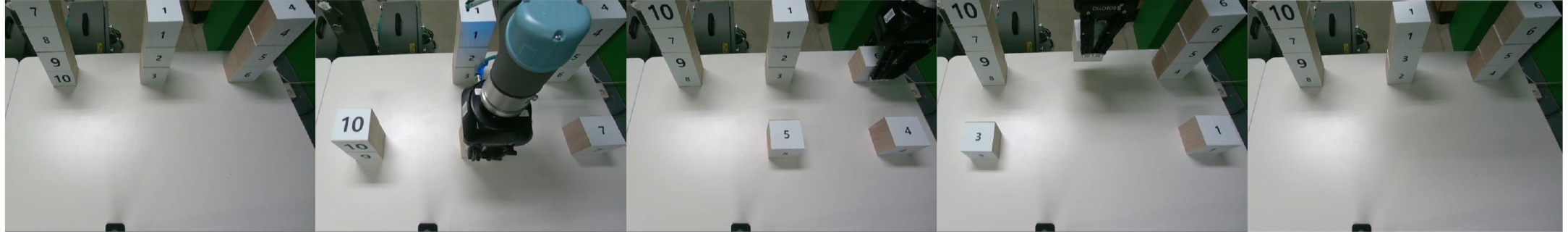}
    \caption{Physical robotic demonstration of our planner on Blocksworld-new domain. Initially, ten blocks, labeled from 1 to 10, are divided into three stacks and placed on the table (leftmost image). The goal is to restack the blocks at the same position in the following order: 10 on 7, 7 on 9, 9 on 8, 1 on 3, 3 on 2, 6 on 5, and 5 on 4 (rightmost image).}
    \label{real}
    \vspace{-0.7em}
\end{figure*}

\begin{figure*}
    \centering
    \includegraphics[width=\textwidth]{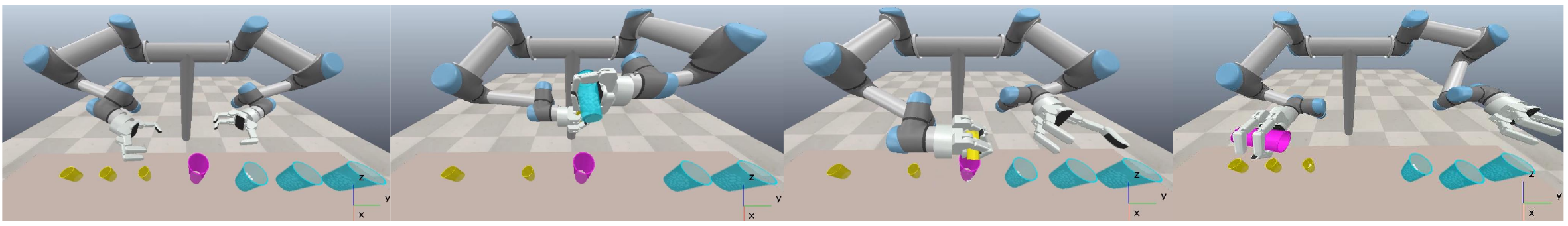}
    \caption{Simulated robotic demonstration of our planner on Barman-new domain. Initially, three ingredients, three shots, and a shaker are placed on the table (leftmost image). The goal is to make a cocktail and pour it into a shot (rightmost image).}
    \label{sim}
    \vspace{-1.3em}
\end{figure*}

\subsection{Performance Analysis}

The success rate and planning time for each experiment are shown in Figure.~\ref{fig.3}. The success rate is verified by the PDDL validator VAL \cite{howey2004val}. The planning time includes the subgoal generation and planning time. For each task, we compared four methods:
\begin{enumerate}
    \item \textbf{CoT planner}: baseline LLM planner which uses chain-of-thought few-shot prompting \cite{wei2022chain,brown2020language} with two or three in-context examples to directly generate a plan with LLM \cite{silver2022pddl,silver2024generalized,zhou2024isr}.
    \item \textbf{FD planner}: baseline symbolic planner using the Fast Downward planner with the "seq-opt-fdss-1" configuration. 
    \item \textbf{Symbolic LLM planner}: our method using the symbolic planner as a subgoal planner, explained in Sec.~\ref{sec.4.A}.
    \item \textbf{MCTS LLM planner}: our method using the MCTS planner as a subgoal planner, explained in Sec.~\ref{sec.4.B}.
\end{enumerate}

{\bf Comparisons:} The CoT planner is the fastest among the four but has the lowest accuracy, with its success rate approaching nearly zero as $n$ increases; on the other hand, the FD planner maintains a 100\% success rate, but its planning time increases exponentially as $n$ grows. This indicates that both baseline methods struggle to solve long-sequential problems in highly complex search spaces. 
In contrast, our Symbolic LLM planner consistently achieved a success rate of 100\%, and our MCTS planner obtained on average $98.5\%, 92.6\%, 88.2\%$ success rates for Barman-new, Blocksworld-new, and Gripper-new domains, respectively.
Compared to the CoT Planner, on average, the planning times of our symbolic and MCTS planners are $6.5\times$/$3.8\times$ (Barman-new), $4.9\times$/$10.2\times$ (Blocksworld-new), and $3.36\times$/$8\times$ (Gripper-new) slower.

It is difficult to compare the performance of our method against other state-of-the-art, LLM-based methods since they use different LLMs or generate non-deterministic results. However, one can estimate comparisons based on the original authors' report. \cite{liu2023llm+} show very low success rates (almost zero) for complex benchmarks like ours, \cite{silver2024generalized} show similar success rates like ours for the single robot benchmark whereas ours is multiple, more complex setup, and \cite{zhou2024isr} show slightly inferior success rates than ours for the Blocksworld when $n\le4$, but it is unclear how it would perform for $n>4$. 
Even though this comparative study is not purely experimental, one can say that the performance of our methods is substantially better than the existing methods.

% 수정 전
% {\bf Symbolic LLM vs. MCTS LLM:} In the Barman-new domain, where the MDL for each sub-problem (making each cocktail) is long, and the domain's state space $\mathcal{S}$ is extensive, the planning time for the Symbolic LLM Planner increases rapidly as $n$ grows. In contrast, the MCTS LLM planner exhibits an almost linear increase in planning time in terms of $n$, resulting in faster performance than the Symbolic LLM planner.
% However, in the Blocksworld-new and Gripper-new domains, the planning time for the Symbolic LLM planner does not increase as quickly as in the Barman-new domain, and it was faster than the MCTS LLM planner, probably because those domains are less complex than the Barman-new domain, and the MDL between subgoals is moderate.
% 수정
{\bf Symbolic LLM vs. MCTS LLM:} In the Barman-new domain, where each sub-task (making a cocktail) requires a long MDL and the domain's state space $\mathcal{S}$ is large, the planning time for the Symbolic LLM planner increases rapidly as $n$ grows. In contrast, the MCTS LLM planner shows an almost linear growth in planning time with respect to $n$, resulting in faster performance than the Symbolic LLM planner.
However, in the Blocksworld-new and Gripper-new domains, the planning time for the Symbolic LLM planner does not increase as quickly as in the Barman-new domain, and it was faster than the MCTS LLM planner. This is probably because these domains are less complex than the Barman-new domain, with a shorter MDL between subgoals.

{\bf Sampled Plans:} We performed further experiments on the number of sampled plans used by the MCTS LLM planner by varying $3\le n_s \le 5$, and observed a general trend of higher success rates, accompanied by an increase in planning time. 
% However, as noted in \cite{hu2023tree}, success rate improvement is limited when $n_s$ exceeds a certain point due to the upper bound on search space complexity, with subgoal decomposition further restricting the space in our case.
However, as noted in \cite{hu2023tree}, success rate improvement is limited when $n_s$ exceeds a certain point due to the upper bound on search space complexity, with subgoal decomposition further restricting the space in our case.

{\bf Ablation Study on Goal Decomposition:} 
We conducted an ablation study on the effectiveness of goal decomposition. We executed our MCTS LLM planner with $n_s = 5$ with and without goal decompositions.  As shown in Fig.~\ref{fig.3}, our planner with goal decomposition achieved a much higher success rate than the one without it, whereas the planner without goal decomposition approached zero success rates for complex problems.

\subsection{Robot Demonstration}
We conducted planning experiments with a real robot in the Blocksworld-new domain to demonstrate the practicality of our neuro-symbolic robot task planners. For the real robot demonstration, we used dual UR5e manipulators with Robotiq 3F grippers. An Intel RealSense D455 RGBD camera was employed for visual input, fixed above the table for a top-down view.
For the Barman-new domain, we conducted experiments in the CoppeliaSim environment \cite{rohmer2013v}, which was set up similarly to the real robot setup. For both experiments, our task planners were integrated into the MoveIt motion planner\cite{chitta2012moveit} in ROS via the translated action primitives.
Key robot action primitives, such as \texttt{pick} and \texttt{place}, were predefined using MoveIt, and task planning results were converted into code composed of these action primitives using the LLM. Once executed, corresponding robot actions were carried out accordingly. 

% 수정: 내용 추가
However, we assume that every high-level action primitive has a feasible low-level solution without explicitly handling motion planning or execution failures. This simplification may lead to uncertainties in real-world execution, affecting execution robustness. Addressing these issues in future work could improve overall system reliability.

% 수정: 섹션 추가
\subsection{Failure Analysis}
% In both real-world and simulation experiments, we observed two main types of failures: execution errors and planning errors. Execution errors were mainly caused by stability issues, such as failed grasps or collapsed stacks due to unstable placement, as well as occlusion in cluttered environments causing inaccurate planning formulation. Planning errors were more common in the MCTS LLM planner. In the Blocksworld-new domain, the planner struggled with spatial reasoning, often misordering block stacking sequences. In the Gripper-new domain, it occasionally executed unnecessary actions, such as moving irrelevant balls, indicating misinterpretation of the goal state.
In both real-world and simulation experiments, failures fell into two categories: execution and planning. Execution failures mostly stemmed from stability issues, such as occlusion in cluttered environments leading to inaccurate planning formulation, failed grasps, or collapsed stacks of blocks. Planning failures were more common in the MCTS LLM planner. In the Blocksworld-new domain, it struggled with spatial reasoning and misordered block stacking sequences, while in the Gripper-new domain, it misinterpreted the goal state, occasionally moving irrelevant balls.

%%%%%%%%%%%%%%%%%%%%%%%%%%%%%%%%%%%%%%%%%%%%%%%%%%%%%%%%%%%%%%%%%%%%%%%%%%%%%%%%

%%%%%%%%%%%%%%%%%%%%%%%%%%%%%%%%%%%%%%%%%%%%%%%%%%%%%%%%%%%%%%%%%%%%%%%%%%%%%%%%

\section{CONCLUSION AND FUTURE WORK} \label{sec:conclusion}
This paper proposes a novel task-planning method based on neuro-symbolic language models by decomposing a complicated, long-sequential goal into multi-level subgoals.
Our planner performs much faster than the baseline symbolic methods, achieving high accuracy. 
% We would like to pursue a couple of future directions to improve our current method. 
% The criterion for choosing the level of goal decompositions and picking either a symbolic or MCTS planner for the subgoal is empirical. More automated and thorough strategies for this decision problem are needed.
% We also need more investigation into integrating our tasking planning pipeline to motion planning, \ie the TAMP. 
Future improvements include developing automated strategies for selecting the level of goal decomposition and choosing between symbolic and MCTS planners, which currently rely on empirical criteria. Further integration of our task-planning pipeline with motion planning (\ie the TAMP) is also needed.
% 수정: 내용 추가
Moreover, ensuring the generalizability of the subgoal decomposition in more complex real-world tasks remains essential.
Expanding evaluations to diverse environments, such as multi-agent systems, will further enhance their robustness and generalizability

%%%%%%%%%%%%%%%%%%%%%%%%%%%%%%%%%%%%%%%%%%%%%%%%%%%%%%%%%%%%%%%%%%%%%%%%%%%%%%%%


%\section*{APPENDIX}

\section*{ACKNOWLEDGMENT}
This work was supported in part by the ITRC/IITP Program (IITP-2025-RS-2020-II201460), and in part by the NRF (NRF-2022R1A2B5B03001385) in South Korea.

% \bibliographystyle{ieeetr}
% \bibliography{main}
\bibliographystyle{IEEEtran}
\bibliography{IEEEabrv,main}

\end{document}